\documentclass{new_tlp}
\usepackage{times,helvet,courier}

\usepackage{url}
\usepackage{amsmath}
\let\proof\relax

\usepackage{amssymb,amsthm,mathrsfs,amsfonts,dsfont} 
\usepackage{xcolor}
\usepackage{version}
\usepackage{comment}
\usepackage{mdframed}
\usepackage{graphicx}
\usepackage{graphics} 
\usepackage{epsfig} 
\usepackage{multirow}
\usepackage{xspace}
\usepackage{color}
\usepackage{asplisting}
\definecolor{darkgreen}{RGB}{0,60,0}
\definecolor{darkgray}{RGB}{80,80,80}

\lstnewenvironment{asp}
{
	\lstset{
		language=asp,
		showstringspaces=false,
		formfeed=\newpage,
		tabsize=4,
		commentstyle=\color{darkgreen},
		basicstyle=\ttfamily\footnotesize,
		keywordstyle=\color{blue},
		numbers=none,
		breaklines=true,
		literate={~} {$\sim$}{1},
	}
}
{
}
 
\theoremstyle{definition}

\theoremstyle{remark}

\newcommand{\derives}{\ensuremath{\mbox{\,$:$--}\,}\xspace}

\newcommand{\naf}{\ensuremath{not\ }}
\newcommand{\clingo}{\textsc{clingo}\xspace}

\title[Operating Room (Re)Scheduling with Bed Management via ASP]{Operating Room (Re)Scheduling with \\Bed Management via ASP}

 \author[C. Dodaro et al.]
 {CARMINE DODARO\\
 University of Calabria, Italy\\
 \email{dodaro@mat.unical.it}
 \and
  GIUSEPPE GALAT\`A\\
 SurgiQ srl, Italy\\
 \email{giuseppe.galata@surgiq.com}
 \and
 MUHAMMAD KAMRAN KHAN, MARCO MARATEA\\
 University of Genoa, Italy\\
 \email{muhammad.kamrankhan@edu.unige.it,marco.maratea@unige.it}
 \and 
 IVAN PORRO\\
 SurgiQ srl, Italy\\
\email{ivan.porro@surgiq.com}
 }

 \jdate{March 2003}
\pubyear{2003}
\pagerange{\pageref{firstpage}--\pageref{lastpage}}
\doi{S1471068401001193}

\begin{document}

\label{firstpage}
\maketitle
%


%
%
 

\begin{abstract}
  The Operating Room Scheduling (ORS) problem is the task of assigning patients to operating rooms, taking into account different specialties, lengths and priority scores of each planned surgery, operating room session durations, and the availability of beds for the entire length of stay both in the Intensive Care Unit and in the wards.
  A proper solution to the ORS problem is of primary importance for the healthcare service quality and the satisfaction of patients in hospital environments.
  In this paper we first present a solution to the problem based on Answer Set Programming (ASP). The solution is tested on benchmarks with realistic sizes and parameters, on three scenarios for the target length on 5-day scheduling, common in small-medium sized hospitals, and results show that ASP is a suitable solving methodology for the ORS problem in such setting. Then, we also performed a scalability analysis on the schedule length up to 15 days, which still shows the suitability of our solution also on longer plan horizons. Moreover, we also present an ASP solution for the rescheduling problem, i.e. when the off-line schedule cannot be completed for some reason. Finally, we introduce a web framework for managing ORS problems via ASP that allows a user to insert the main parameters of the problem, solve a specific instance, and show results graphically in real-time. Under consideration in Theory and Practice of Logic Programming (TPLP).\footnote{This paper is an extended and revised version of a conference paper appearing in the proceedings of the 3rd International Joint Conference on Rules and Reasoning (RuleML+RR 2019) ~\cite{DBLP:conf/ruleml/DodaroGKMP19}.}\footnote{{\bf Disclaimer:} two of the authors
of this paper, Ivan Porro and Giuseppe Galat\`a, have business interest in SurgiQ.}
  


\end{abstract}

\section{Introduction}

The Operating Room Scheduling (ORS) \cite{abedini_operating_2016,aringhieri_two_2015,landa_hybrid_2016,molina-pariente_new_2015} problem is the task of assigning patients to operating rooms, taking into account different specialties, surgery durations, and the availability of beds for the entire length of stay (LOS) both in the Intensive Care Unit (ICU) and in the wards.  Given that patients may have priorities, the solution has to find an accommodation for the patients with highest priorities, and then to the other with lower priorities, if space is still available, at the same time taking into proper account bed availability. A proper solution to the ORS problem is crucial for improving the whole quality of the healthcare and the satisfaction of patients. Indeed, modern hospitals are often characterized by long surgical waiting lists, which are caused by inefficiencies in operating room planning, leading to an obvious dissatisfaction of patients. Complex combinatorial problems, possibly involving optimizations, such as the ORS problem, are usually the target applications of knowledge representation and reasoning formalisms such as Answer Set Programming (ASP)~\cite{DBLP:conf/iclp/GelfondL88,DBLP:journals/ngc/GelfondL91,DBLP:journals/amai/Niemela99,baral2003,DBLP:journals/cacm/BrewkaET11}, which is particularly suited for applications given its simple but rich syntax~\cite{aspcore2}, its intuitive semantics, combined with the readability of specifications (always appreciated by users), and the availability of efficient solvers, e.g. \clingo~\cite{DBLP:journals/ai/GebserKS12} and {\sc wasp}~\cite{AlvianoADLMR19}. 



In this paper we present a solution to the problem based on Answer Set Programming (ASP). In such solution, problem specifications are modularly added as ASP rules to compose the ASP encoding, and then \clingo is used to solve the resulting ASP program. The solution is tested on benchmarks with realistic sizes and parameters, for the target length on 5-day scheduling, common in small-medium sized hospitals, on three scenario: a first scenario is characterized by an abundance of available beds, so that the constraining resource becomes the OR time, while in the second and third scenario, corresponding to mild or severe bed shortage, the number of beds is the constrained resource. Testing our algorithm also in such extreme situation is important because the combination of the dwindling number of available beds and the increase in older population is causing increasing strain on the national health systems of most developed countries. For example, between Q1 2010/11 and Q3 2018/19, the total number of NHS hospital beds decreased by 12\%, from 144,455 to 127,589 \footnote{https://www.nuffieldtrust.org.uk/resource/hospital-bed-occupancy}. The latest official NHS data show\footnote{See data stored on https://www.england.nhs.uk/statistics/statistical-work-areas/bed-availability-and-occupancy/bed-data-overnight/} that in Q2 2019/20 (from July to September 2019) 70 out of 202 Trusts had an average bed occupancy $>90\%$. Hospitals cannot operate at or close at 100\% occupancy, as spare bed capacity is needed to accommodate variations in demand and ensure that patients can flow through the system. 

Overall, results show that ASP is a suitable solving methodology on all considered scenarios, given that our solution is able to utilize efficiently whichever resource is more constrained; moreover, this is obtained in short timings in line with the needs of the application. Then, we also performed a scalability analysis on the schedule length for all considered scenarios, up to three times the target planning length: results show that our solution is still able to utilize efficiently whichever resource is more constrained even on longer planning horizons, with some degradation only appearing in some cases for the longer planning length. Moreover, we also present an ASP solution for the rescheduling problem, i.e. when the off-line schedule cannot be completed for some reason. Finally, we introduce a web framework for managing ORS problems via ASP that allows a user to insert the main parameters of the problem, solve a specific instance, and show results graphically in real-time, in such a way being able to use our solution without installing nothing specific on the user laptop.

To summarize, the main contributions of this paper are the following:
\begin{itemize}
	\item We provide an ASP encoding for solving the ORS problem (Section~\ref{sec:models}).
	\item We run an experimental analysis assessing the good performance of our ASP solution both for the target 5-day scheduling length (Section~\ref{sec:experiments}.2) and for the scalability analysis w.r.t. scheduling length (Section~\ref{sec:experiments}.3).
	\item We provide an ASP encoding and an experimental analysis for the rescheduling problem (Section~\ref{sec:resch}).
	\item We describe a Graphical User Interface which uses our ASP solution to produce a real-time scheduling of operating rooms (Section~\ref{sec:appl}).
\end{itemize}

The paper is completed by Section~\ref{sec:back}, which contains needed preliminaries about ASP, by an informal description of the ORS problem in
Section~\ref{sec:spec}, by the related work analysis in Section~\ref{sec:rel}, and  by conclusions and possible topics for future research in Section~\ref{sec:conc}.

\section{Background on ASP}
\label{sec:back}

Answer Set Programming (ASP) \cite{DBLP:journals/cacm/BrewkaET11} is a programming paradigm
developed in the field of nonmonotonic reasoning and logic programming.
In this section, we overview the language of ASP. 
More detailed descriptions and a more formal account of ASP, including
the features of the language employed in this paper, can be found
in~\cite{DBLP:journals/cacm/BrewkaET11,DBLP:journals/tplp/CalimeriFGIKKLM20,DBLP:journals/tplp/GebserHKLS15,DBLP:journals/tplp/AlvianoFG15}.
Hereafter, we assume the reader is familiar with logic programming conventions.

\paragraph{Syntax.}
The syntax of ASP is similar to the one of Prolog.
Variables are strings starting with uppercase letter and
constants are non-negative integers or strings starting with lowercase letters.
A {\em term} is either a variable or a constant.
A {\em standard atom} is an expression $p(t_1, \ldots, t_n)$, where $p$ is a
{\em predicate} of arity $n$ and $t_1, \ldots, t_n$ are terms.
An atom $p(t_1, \ldots, t_n)$ is ground if $t_1, \ldots, t_n$ are constants.
A {\em ground set} is a set of pairs of the form $\langle consts\! :\!conj \rangle$,
where $consts$ is a list of constants and $conj$ is a conjunction of ground standard atoms.
A {\em symbolic set} is a set specified syntactically as
$\{Terms_1 : Conj_1; \cdots; Terms_t : Conj_t \}$,
where $t>0$, and for all $i \in [1,t]$, each $Terms_i$ is a list of terms such that $|Terms_i| = k > 0$, and  each $Conj_i$ is a conjunction of standard atoms.
A {\em set term} is either a symbolic set or a ground set.
Intuitively, a set term $\{X\! :\! a(X,c), p(X);Y\! :\! b(Y,m)\}$
stands for the union of two sets: the first one contains the $X$-values making the conjunction $a(X,c), p(X)$ true, and the second one contains the $Y$-values making the conjunction $b(Y,m)$ true.
An {\em aggregate function} is of the form $f(S)$, where $S$ is a
set term, and $f$ is an {\em aggregate function symbol}.
%
%
Basically, aggregate functions map multisets of constants to a constant.
The most common functions implemented in ASP systems are the following:

\begin{itemize}	
	\item {\textit{\#count}}, number of terms;
	
	\item {\textit{\#sum}}, sum of integers.
	
\end{itemize}
An {\em aggregate atom} is of the form $f(S) \prec T$, where $f(S)$ is an
aggregate function, $\prec\ \in \{<, \leq, >, \geq, \neq, =\}$
is a comparison operator, and $T$ is a term called guard.
An aggregate atom $f(S) \prec T$ is ground if $T$ is a constant and
$S$ is a ground set.
An \emph{atom} is either a standard atom or an aggregate atom.
A {\em rule} $r$ has the following form:

\begin{center}
	$a_1 \ \vee \ \ldots \ \vee \ a_n \ \derives \ b_1,\ldots, b_k, \naf b_{k+1},\ldots, \naf b_m.$
\end{center}

\noindent where $a_1,\ldots ,a_n$ are standard atoms, $b_1,\ldots ,b_k$ are atoms,
$b_{k+1},\ldots ,b_m$ are standard atoms, and $n,k,m\geq 0$.
A literal is either a standard atom $a$ or its negation $\naf a$.
The disjunction $a_1 \vee \ldots  \vee a_n$ is the {\em head} of $r$, while
the conjunction $b_1 , \ldots,  b_k, \naf b_{k+1} ,$ $\ldots, \naf b_m$ is its {\em body}. Rules with empty body are called {\em facts}.
Rules with empty head are called {\em constraints}. 
A variable that appears uniquely in set terms of a rule $r$ is said to be {\em local} in $r$, otherwise it is a {\em global} variable of $r$.
An ASP program is a set of {\em safe} rules, where
a rule $r$ is {\em safe} if the following conditions hold:
{\em (i)} for each global variable $X$ of $r$ there is a positive standard atom
$\ell$ in the body of $r$ such that $X$ appears in $\ell$; and
{\em (ii)} each local variable of $r$ appearing in a symbolic set
$\{ \mathit{Terms}\! :\! \mathit{Conj}\}$ also appears in a positive atom in $\mathit{Conj}$.

A {\em weak constraint}~\cite{DBLP:journals/tkde/BuccafurriLR00} $\omega$ is of the form:

\begin{center}
	$:\sim b_1,\ldots, b_k, \naf b_{k+1},\ldots, \naf b_m.\ [w@l] $
\end{center}

\noindent where $w$ and $l$ are the weight and level of $\omega$, respectively.
(Intuitively, $[w@l]$ is read ``as weight $w$ at level $l$'', 
where weight is the ``cost'' of violating the condition in the body, 
whereas levels can be specified for defining a priority among preference criteria).
An ASP program with weak constraints is $\Pi = \langle P,W \rangle$, where $P$ is a program
and $W$ is a set of weak constraints.

A standard atom, a literal, a rule, a program or a weak constraint is {\em ground} if no variables appear in it.

\paragraph{Semantics.}
Let $P$ be an ASP program. The {\em Herbrand universe} $U_{P}$ and
the {\em Herbrand base} $B _{P}$ of $P$ are defined as usual.
%
The ground instantiation $G_P$ of $P$ is the set of all the ground instances of rules of $P$ that can be obtained by substituting variables with constants from $U_{P}$.

An {\em interpretation} $I$ for $P$ is a subset $I$ of $B_{P}$.
A ground literal $\ell$ (resp., $\naf \ell$) is true w.r.t. $I$
if $\ell \in I$ (resp., $\ell \not\in I$), and false otherwise.
An aggregate atom is true w.r.t. $I$ if the evaluation of its aggregate function
(i.e. the result of the application of $f$ on the multiset $S$) with respect to $I$
satisfies the guard; otherwise, it is false.

A ground rule $r$ is {\em satisfied} by $I$
if at least one atom in the head is true w.r.t. $I$ whenever all conjuncts of the body
of $r$ are true w.r.t. $I$.

A model is an interpretation that satisfies all rules of a program.
Given a ground program  $G_P$ and an interpretation  $I$, the
{\em reduct} of $G_P$ w.r.t. $I$ is the subset $G_P^I$ of $G_P$ obtained
by deleting from $G_P$ the rules in which a body literal is false w.r.t. $I$.
An interpretation $I$ for $P$ is an {\em answer set} (or stable model)
for  $P$ if $I$ is a minimal model (under subset inclusion) of $G_P^I$
(i.e.  $I$ is a minimal model for $G_P^I$)~\cite{DBLP:journals/ai/FaberPL11,DBLP:journals/tocl/Ferraris11}.
For a detailed discussion on the semantics of ASP programs with aggregates, we refer the reader to \cite{DBLP:journals/tplp/AlvianoFG15}.

Given a program with weak constraints $\Pi = \langle P,W \rangle$, the semantics of $\Pi$ extends from the basic case defined above. Thus, let $G_{\Pi} = \langle G_P,G_W \rangle$ be the instantiation  of $\Pi$; a constraint $\omega \in G_W$ is violated by an interpretation $I$ if all the literals in $\omega$ are true w.r.t. $I$.
An {\em optimum answer set} for $\Pi$ is an answer set of $G_P$  that minimizes
the sum of the weights of the violated weak constraints in $G_W$ in a prioritized way.

\paragraph{Syntactic shortcuts.}
In the following, we also use \textit{choice rules} of the form $\{p\}$, where $p$ is an atom. Choice rules can be viewed as a syntactic shortcut for the rule $p \vee p'$, where $p'$ is a fresh new standard atom not appearing elsewhere in the program.

\section{Problem Description}
\label{sec:spec}
In this section we provide an informal description of the ORS problem and its requirements.
As we already said in the introduction, 
most modern hospitals are characterized by a very long surgical waiting list, often worsened, if not altogether caused, by inefficiencies in operating room planning. A very important factor is represented by the availability of beds in the wards and, if necessary, in the ICU for each patient for the entire duration of their stay.
This means that hospital planners have to balance the need to use the OR time with the maximum efficiency with an often reduced bed availability.  

In this paper, the elements of the waiting list are called \textit{registrations}. Each registration links a particular surgical procedure, with a predicted surgery duration and length of stay in the ward and in the ICU, to a patient.
The overall goal of the ORS problem is to assign the maximum number of registrations to the operating rooms (ORs), taking into account the availability of beds in the associated wards and in the ICU. This approach entails that the resource optimized is the one, between the OR time and the beds, that represents the bottleneck in the particular scenario analyzed. 

As first requirement of the ORS problem, the assignments must guarantee that the sum of the predicted duration of surgeries assigned to a particular OR session does not exceed the length of the session itself: this is referred in the following as \textit{surgery requirement}. 
Moreover, registrations are not all equal: they can be related to different medical conditions and can be in the waiting list for different periods of time.
These two factors are unified in one concept: \textit{priority}.
Registrations are classified according to three different priority categories, namely $P_{1}$, $P_{2}$ and $P_{3}$.
The first one gathers either very urgent registrations or the ones that have been in the waiting list for a long period of time; it is required that these registrations are all assigned to an OR.
Then, the registrations of the other two categories are assigned to the top of the ORs capacity, prioritizing the $P_{2}$ over the $P_{3}$ ones (\textit{minimization}).

Regarding the bed management part of the problem, we have to ensure that a registration can be assigned to an OR only if there is a bed available for the patient for the entire LOS. In particular, we have considered the situation where each specialty is related to a ward with a variable number of available beds exclusively dedicated to the patients associated to the specialty.
This is referred in the following as \textit{ward bed requirement}. The ICU is a particular type of ward that is accessible to patients from any specialty. However, only a small percentage of patients is expected to need to stay in the ICU. This requirement will be referred as the \textit{ICU bed requirement}. Obviously, during their stay in the ICU, the patient does not occupy a bed in the specialty's ward.  

In our model, a patient's LOS has been divided in the following phases:
\begin{itemize}
    \item a LOS in the ward before surgery, in case the admission is programmed a day (or more) before the surgery takes place; and
    \item the LOS after surgery, which can be further subdivided into the ICU LOS and the following ward LOS.
\end{itemize}



The encoding described in Section~\ref{sec:models} supports the generation of an optimized schedule of the surgeries either in the case where the bottleneck is represented by the OR time or by the bed availability.  

\section{ASP Encoding for the ORS problem}
\label{sec:models}
\begin{figure*}[t!]
\begin{asp}
$(r_{1})$$\phantom{_0}$ {x(R,P,O,S,D)} :- registration(R,P,_,_,SP,_,_), mss(O,S,SP,D).
$(r_{2})$$\phantom{_0}$ :- x(R,P,O,S1,_), x(R,P,O,S2,_), S1 != S2.
$(r_{3})$$\phantom{_0}$ :- x(R,P,O1,S,_), x(R,P,O2,S,_), O1 != O2.
$(r_{4})$$\phantom{_0}$ surgery(R,SU,O,S) :- x(R,_,O,S,_), registration(R,_,SU,_,_,_,_).
$(r_{5})$$\phantom{_0}$ :- x(_,_,O,S,_), duration(N,O,S), #sum{SU,R:surgery(R,SU,O,S)}>N.
$(r_{6})$$\phantom{_0}$ stay(R,(D-A)..(D-1),SP) :- registration(R,_,_,LOS,SP,_,A), 
$\qquad\qquad\qquad$x(R,_,_,_,D), A>0.
$(r_{7})$$\phantom{_0}$ stay(R,(D+ICU)..(D+LOS-1),SP) :- registration(R,_,_,LOS,SP,ICU,_), 
$\qquad\qquad\qquad$x(R,_,_,_,D), LOS>ICU.
$(r_{8})$$\phantom{_0}$ stayICU(R,D..(D+ICU-1)) :- registration(R,_,_,_,_,ICU,_), 
$\qquad\qquad\qquad$x(R,_,_,_,D), ICU>0.
$(r_{9})$$\phantom{_0}$ :- #count {R: stay(R,D,SP)}>AV, SP>0, beds(SP,AV,D).
$(r_{10})$ :- #count {R: stayICU(R,D)}>AV, beds(0,AV,D).
$(r_{11})$ :- N = totRegsP1 - #count{R: x(R,1,_,_,_)}, N > 0.
$(r_{12})$ :~ N = totRegsP2 - #count{R: x(R,2,_,_,_)}. [N@3]
$(r_{13})$ :~ N = totRegsP3 - #count{R: x(R,3,_,_,_)}. [N@2]     
\end{asp}
    \caption{ASP encoding of the ORS problem.}
    \label{fig:encoding1}
\end{figure*}

Starting from the specifications in the previous section,
here the ASP encoding of the ORS scheduling problem is described in the ASP language, in particular following the input language of {\sc clingo}.
It is important to emphasize here that, albeit \textsc{clingo} is compliant with the ASP-Core2~\cite{aspcore2} input language, it supports a richer syntax and slightly different semantics, see \cite{DBLP:journals/tplp/GebserHKLS15} for a formal description of the language.
Next two sub-sections present the data model and the encoding  itsself, respectively. 

\subsection{Data Model}\label{sec:datamodelors}
The input data is specified by means of the following atoms:
\begin{itemize}
	\item Instances of \textit{registration(R,P,SU,LOS,SP,ICU,A)} represent the registrations, characterized by an id ($R$), a priority score ($P$), a surgery duration ($SU$) in minutes, the overall length of stay both in the ward and the ICU after the surgery ($LOS$) in days, the id of the specialty ($SP$) it belongs to, a length of stay in the ICU ($ICU$) in days, and finally a parameter representing the number of days in advance ($A$) the patient is admitted to the ward before the surgery. 
	\item Instances of \textit{mss(O,S,SP,D)} link each operating room ($O$) to a session ($S$) for each specialty ($SP$) and planning day ($D$) as established by the hospital Master Surgical Schedule (MSS).
	\item The OR sessions are represented by the instances of the predicate \textit{duration(N,O,S)}, where $N$ is the session duration.
	\item
	Instances of \textit{beds(SP,AV,D)} represent the number of available beds ($AV$) for the beds associated to the specialty $SP$ in the day $D$. The ICU is represented by giving the value $0$ to $SP$.
\end{itemize}

The output is an assignment represented by atoms of the form \textit{x(R,P,O,S,D)}, where the intuitive meaning is that the registration $R$ with priority $P$ is assigned to the OR $O$ during the session $S$ and the day $D$. It is important to emphasize here that the priority $P$ is not actually needed in the output, however it is included because it improves the readability of the encoding presented in the subsequent section.

\subsection{Encoding}\label{sec:encodingors}

The related encoding is shown in Figure~\ref{fig:encoding1}, and is described in the following.
Rule ($r_1$) guesses an assignment for the registrations to an OR in a given day and session among the ones permitted by the MSS for the particular specialty the registration belongs to.

The same registration should not be assigned more than once, in different OR or sessions. This is assured by constraints ($r_2$) and ($r_3$). Note that in our setting there is no requirement that every registration must actually be assigned.

\paragraph{Surgery requirement.}
With rules ($r_4$) and ($r_5$) we impose that the total length of surgery durations assigned to a session is less than or equal to the session duration.

Rules ($r_6$)-($r_{10}$) deal with the presence and management of beds. 
In particular, rule ($r_6$) assigns a bed in the ward to each registration assigned to an OR, for the days before the surgery. Rule ($r_{7}$) assigns a ward bed for the period after the patient was dismissed from the ICU and transferred to the ward. Rule ($r_{8}$) assigns a bed in the ICU.
\paragraph{Ward bed requirement.}
Rule ($r_{9}$) ensures that the number of patients occupying a bed in each ward for each day is never larger than the number of available beds.
\paragraph{ICU bed requirement.}
Finally, rule ($r_{10}$) performs a similar check as the one in rule ($r_{9}$), but for the ICU.

\paragraph{Minimization.}
We remind that we want to be sure that every registration having priority $1$ is assigned, then we assign as much as possible of the others, giving precedence to registrations having priority $2$ over those having priority $3$. This is accomplished through constraint ($r_{11}$) for priority $1$ and the weak constraints ($r_{12}$) and ($r_{13}$) for priority 2 and 3, respectively, where \textit{totRegsP1}, \textit{totRegsP2}, and \textit{totRegsP3} are constants representing the total number of registrations having priority $1$, $2$ and $3$, respectively.

Minimizing the number of unassigned registrations could cause an implicit preference towards the assignments of the registrations with shorter surgery durations. 
To avoid this effect, one can consider to minimize the idle time; however, this is in general slower from a computational point of view and often unnecessary, since the  preference towards shorter surgeries is already mitigated by our three-tiered priority schema. 

\paragraph{\bf Remark.} We note that, given that the MSS is fixed, our problem and encoding could be decomposed by considering each specialty separately in case the beds are not a constrained resource, as will be the case for one of our scenario. We decided not to use this property because $(i)$ this is the description of a practical application that is expected to be extended over time and to correctly work even if the problem becomes non-decomposable, e.g. a (simple but significant) extension in which a room is shared among specialties leads to a problem which is not anymore decomposable, and $(ii)$ it is not applicable to all of our scenario. Additionally, even not considering this property at the level of the encoding, the experimental analysis that we will present is already satisfactory for our use case even when the decomposition could be applied.

\section{Experimental Results for Scheduling}\label{sec:experiments}

In this section we report about the results of an empirical analysis of the ORS encoding. Data have been randomly generated but having parameters and sizes inspired by real data. Experiments were run on a Intel Core i7-7500U CPU @ 2.70GHz with 7.6 GB of physical RAM. The ASP system used was \clingo~\cite{DBLP:conf/iclp/GebserKKOSW16}, version 5.5.2, with the "$--$restart$-$on$-$model" option enabled. 

\subsection{ORS benchmarks}\label{subsec:schedulerresults}
The employed encoding is composed by the ASP rules $(r_1), \dots, (r_{13})$ from Figure~\ref{fig:encoding1}.
The test cases we have assembled are based on the requirements of a typical small-medium size Italian hospital, with five surgical specialties to be managed over the widely used 5-day planning period. Three different scenarios were assembled. The first one (scenario A) is characterized by an abundance of available beds, so that the constraining resource becomes the OR time. For the second one (scenario B), we reduced the number of beds, in order to test the encoding in a situation with plenty of OR time but few available beds. Scenario B is pushed further in scenario C, where the number of beds is further reduced, to test our encoding also in this extreme situation. Each scenario was tested 10 times with different randomly generated inputs.

\begin{table*}[t]
    \caption{Bed availability for each specialty and in each day in scenario A.}
    \centering
	\label{tab:bedsAinput}
	\begin{tabular}{cccccc}
	\hline\hline
		Specialty & Monday & Tuesday & Wednesday & Thursday & Friday  \\ \hline
		0 (ICU) & 40  & 40 & 40 & 40 & 40  \\
		1       & 80  & 80 & 80 & 80 & 80  \\
		2       & 58  & 58 & 58 & 58 & 58  \\
		3       & 65  & 65 & 65 & 65 & 65  \\
		4       & 57  & 57 & 57 & 57 & 57  \\
		5       & 40  & 40 & 40 & 40 & 40  \\
	\hline\hline
	\end{tabular}
\end{table*}

\begin{table*}[t]
    \caption{Bed availability for each specialty and in each day in scenario B.}
	\centering
	\label{tab:bedsBinput}
	\begin{tabular}{cccccc}
	\hline\hline
		Specialty & Monday & Tuesday & Wednesday & Thursday & Friday  \\ \hline
		0 (ICU) & 4   & 4  & 5  & 5  & 6  \\
		1       & 20  & 30 & 40 & 45 & 50  \\
		2       & 10  & 15 & 23 & 30 & 35  \\
		3       & 10  & 14 & 21 & 30 & 35  \\
		4       & 8   & 10 & 14 & 16 & 18  \\
		5       & 10  & 14 & 20 & 23 & 25  \\
	\hline\hline
	\end{tabular}
\end{table*}

\begin{table*}[t]
    \caption{Bed availability for each specialty and in each day in scenario C.}
	\centering
	\label{tab:bedsCinput}
	\begin{tabular}{cccccc}
	\hline\hline
		Specialty & Monday & Tuesday & Wednesday & Thursday & Friday  \\ \hline
		0 (ICU) & 4   & 4  & 5  & 5  & 6  \\
		1       & 10  & 15 & 20 & 25 & 30 \\
		2       & 7   & 10 & 11 & 14 & 18 \\
		3       & 7   & 10 & 13 & 16 & 20  \\
		4       & 4   & 6  & 8  & 11 & 13  \\
		5       & 6   & 9  & 12 & 15 & 18  \\
	\hline\hline
	\end{tabular}
\end{table*}

The characteristics of the tests are the following:
\begin{itemize}
	\item 3 different benchmarks, comprising a planning period of 5 working days, and different numbers of available beds, as reported in Table~\ref{tab:bedsAinput}, Table~\ref{tab:bedsBinput} and Table~\ref{tab:bedsCinput}  for scenario A, B, and C, respectively;
	\item 10 ORs, unevenly distributed among the specialties;
	\item 5 hours long morning and afternoon sessions for each OR, summing up to a total of 500 hours of ORs available time for each benchmark;
	\item 350 generated registrations, from which the scheduler will draw the assignments.
	In this way, we simulate the common situation where a hospital manager takes an ordered, w.r.t. priorities, waiting list and tries to assign as many elements as possible to each OR.
\end{itemize}
The surgery durations have been generated assuming a normal distribution, while the priorities have been generated from a uneven distribution of three possible values (with weights respectively of 0.20, 0.40 and 0.40 for registrations having priority 1, 2 and 3, respectively). The lengths of stay (total LOS after surgery and ICU LOS) have been generated using a truncated normal distribution, in order to avoid values less than 1. In particular for the ICU, only a small percentage of patients have been generated with a predicted LOS while the large majority do not need to pass through the ICU and their value for the ICU LOS is fixed to 0. Finally, since the LOS after surgery includes both the LOS in the wards and in the ICU, the value generated for the ICU LOS must be less than or equal to the total LOS after surgery.

\begin{table*}[h]
    \caption{Parameters for the random generation of the scheduler input.}
	\centering
	\setlength{\tabcolsep}{0.09cm}
	\label{tab:schedinput}
	\begin{tabular}{cccccccc}
	\hline
Specialty & Reg. & ORs & Surgery Duration (min) & LOS (d) & ICU (\%) & ICU LOS (d) & LOS (d) \\
& & & mean (std) &  mean (std) & & mean (std) & before surgery \\
\hline\hline
		1 & 80  & 3 & 124 (59.52) & 7.91 (2) & 10 & 1 (1) & 1 \\
		2 & 70  & 2 & 99  (17.82) & 9.81 (2) & 10 & 1 (1) & 1 \\
		3 & 70  & 2 & 134 (25.46) & 11.06 (3)& 10 & 1 (1) & 1 \\
		4 & 60  & 1 & 95  (19.95) & 6.36 (1) & 10 & 1 (1) & 0 \\
		5 & 70  & 2 & 105 (30.45) & 2.48 (1) & 10 & 1 (1) & 0 \\
	Total & 350 & 10&             & & & & \\ \hline\hline
	\end{tabular}
\end{table*}

 The parameters of the test have been summed up in Table~\ref{tab:schedinput}. In particular, for each specialty (1 to 5), we reported the number of registrations generated, the number of ORs assigned to the specialty, the mean duration of surgeries with its standard deviation, the mean LOS after the surgery with its standard deviation, the percentage of patients that need to stay in the ICU, the mean LOS in the ICU with its standard deviation and, finally, the LOS before the surgery (i.e. the number of days, constant for each specialty, the patient is admitted before the planned surgery is executed).


\subsection{Results}
Results of the experiments are reported for scenario A in Table~\ref{tab:schedresultsA}, for scenario B in Table~\ref{tab:schedresultsB}, and for scenario C in Table~\ref{tab:schedresultsC}. 

\begin{table*}[t]
    \caption{Scheduling results for the scenario A benchmark.}
	\centering
	\label{tab:schedresultsA}
	\begin{tabular}{cccccc}
		\hline\hline
\multicolumn{4}{c}{Assigned Registrations} & & \\
\cline{1-4}\\
Priority 1     & Priority 2     & Priority 3    & Total          & OR Time Eff.     & Bed Occupancy Eff. \\ 
\hline
		62 / 62 & 132 / 150 & 72 / 138 & 266 / 350 & 96.6\% & 52.0\% \\ 
		72 / 72 & 128 / 145   & 64 / 133 & 264 / 350 & 95.6\% & 51.0\%\\ 
		71 / 71 & 132 / 132 & 69 / 147 & 272 / 350 & 96.7\% & 96.7\% \\ 
		66 / 66 & 138 / 142  & 57 / 142 & 261 / 350 & 96.2\% & 50.7\%\\ 
		79 / 79 & 119 / 130  & 67 / 141 & 265 / 350 & 96.0\% & 51.9\%\\ 
		67 / 67   & 131 / 131  & 66 / 152 & 264 / 350 & 96.6\% & 53.8\%\\ 
		66 / 66 & 121 / 132 & 69 / 152 & 256 / 350 & 96.0\% & 49.8\% \\ 
		69 / 69 & 130 / 135 & 68 / 146 & 267 / 350 & 96.8\% & 51.6\% \\ 
		60 / 60 & 139 / 153 & 59 / 137 & 258 / 350 & 96.8\% & 50.8\% \\ 
		68 / 68   & 138 / 142  & 57 / 139 & 263 / 350 & 95.2\% & 51.3\% \\ 
		\hline\hline
	\end{tabular}
\end{table*}

\begin{table*}[t]
    \caption{Scheduling results for the scenario B benchmark.}
	\centering
	\label{tab:schedresultsB}
	\begin{tabular}{cccccc}
		\hline\hline
		\multicolumn{4}{c}{Assigned Registrations} & & \\
		\cline{1-4}\\
Priority 1     & Priority 2     & Priority 3    & Total          & OR Time Eff.     & Bed Occupancy Eff.  
        \\ \hline
		62 / 62 & 106 / 150 & 13 / 138 & 181 / 350 & 66.3\% & 92.7\% \\ 
		72 / 72 & 77 / 145   & 43 / 133 & 192 / 350 & 67.5\%\ & 94.2\%\\ 
		71 / 71 & 80 / 132 & 38 / 147 & 189 / 350 & 68.2\% & 96.1\%\\\ 
		66 / 66 & 81 / 142  & 41 / 142 & 188 / 350 & 71.4\% & 93.4\% \\ 
		79 / 79 & 90 / 130  & 20 / 141 & 189 / 350 & 69.0\% & 94.1\%\\ 
		67 / 67   & 95 / 131  & 25 / 152 & 187 / 350 & 66.5\%\ & 93.9\%\\ 
		66 / 66 & 92 / 132 & 30 / 152 & 188 / 350 & 71.8\% & 94.1\%\\ 
		69 / 69 & 84 / 135 & 36 / 146 & 189 / 350 & 68.7\% & 92.7\%\\ 
		60 / 60 & 91 / 153 & 34 / 137 & 185 / 350 & 69.7\% & 94.1\% \\ 
		68 / 68   & 82 / 142  & 35 / 139 & 185 / 350 & 69.3\% & 95.1\% \\ 
		\hline\hline
	\end{tabular}
\end{table*}

\begin{table*}[ht]
    \caption{Scheduling results for the scenario C benchmark.}
	\centering
	\label{tab:schedresultsC}
	\begin{tabular}{cccccc}
		\hline\hline
\multicolumn{4}{c}{Assigned Registrations} & & \\
\cline{1-4}\\
Priority 1     & Priority 2     & Priority 3    & Total          & OR Time Eff.     & Bed Occupancy Eff.  
                                                           \\\hline
		62 / 62 & 43 / 150 & 12 / 138 & 117 / 350 & 43.1\% & 85.8\%\\ 
		71 / 71 & 41 / 132 & 6 / 147 & 118 / 350 & 42.9\% & 93.2\%\\ 
		66 / 66 & 40 / 142  & 11 / 142 & 117 / 350 & 42.5\% & 92.0\%\\ 
		79 / 79   & 38 / 130  & 7 / 141 & 124 / 350 & 44.0\% & 93.8\%\\  
		67 / 67 & 42 / 131 & 9 / 152 & 118 / 350 & 41.9\% & 89.8\%\\ 
		69 / 69 & 39 / 135 & 13 / 146 & 121 / 350 & 45.3\% & 94.4\% \\ 
		60 / 60   & 48 / 153  & 10 / 137 & 118 / 350 & 45.3\% & 91.2\% \\ 
		68 / 68   & 38 / 143  & 13 / 139 & 119 / 350 & 44.6\% & 91.5\%\\ 
		\hline\hline
	\end{tabular}
\end{table*}

A time limit of 60 seconds was given and each scenario was run 10 times with different input registrations. 
No run manages to reach the optimal solution within the chosen timeout. However, the quality of the solution improves only marginally even if the timeout is extended up to 5 minutes, which is the largest timeout value we have tried in view of a practical use of the program. For this reason, we decided to keep the timeout value at 60 seconds, which makes the program adapt to be used also for quick testing of "what-if" scenarios and simulations.
For each satisfiable instance out of the 10 runs executed, the tables report in the first three columns the number of the assigned registrations out of the generated ones for each priority, and in the remaining two columns a measure of the total time occupied by the assigned registrations as a percentage of the total OR time available (indicated as "OR Time Eff." in the tables) and the ratio between the beds occupied after the planning to the available ones before the planning (labeled as "Bed Occupancy Eff." in the tables). 
As a general observation, these results show that our solution is able to utilize efficiently whichever resource is more constrained: on scenario A, our solution manages to reach a very high efficiency, over $95\%$, in the use of OR time, while in scenario B achieves an efficiency of bed occupancy between $92\%$ and $95\%$, and over $85\%$ even in the extreme case represented by Scenario C. The same set of generated registrations was used in each scenario, so that the differences in the results can be ascribed only to the different bed configurations.
Taking into consideration a practical use of this solution, the user would be able to individuate and quantify the resources that are more constraining and take the appropriate actions. This means that the solution can also be used to test and evaluate "what if" scenarios.

\begin{figure*}[t!]
	\begin{center}
		\begin{tabular}{c}
			\scalebox{0.5}{\includegraphics{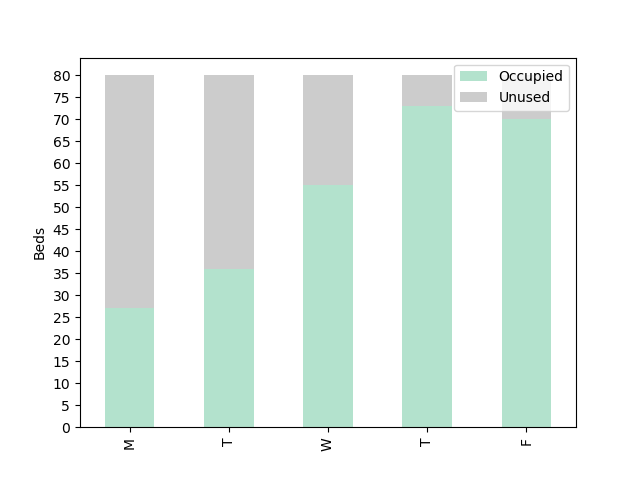}} \\
			\scalebox{0.5}{\includegraphics{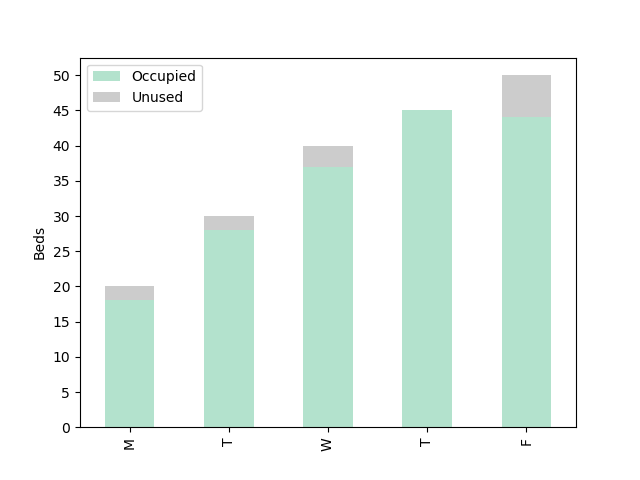}}  \\
			\scalebox{0.5}{\includegraphics{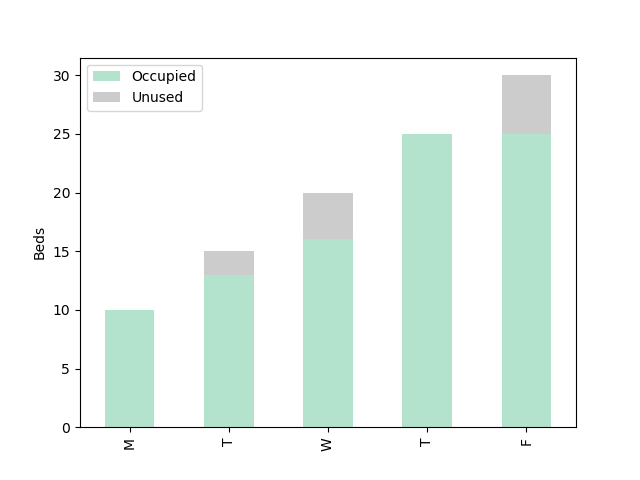}}  
		\end{tabular}
	\end{center}
	\caption{\label{fig:plot} Example of bed occupancy of the ward corresponding to specialty 1 for 5-day scheduling. The plot at the top corresponds to the first instance of scenario A, the one in the middle to the first instance of scenario B. Finally, the one at the bottom corresponds to the first instance of scenario C.}
\end{figure*}

Finally, in Figure~\ref{fig:plot} we (partially) present the results achieved on one instance (i.e. the first instance of Table~\ref{tab:schedresultsA}, Table~\ref{tab:schedresultsB}, and Table~\ref{tab:schedresultsC}) with 350 registrations for 5 days. Each bar represents the total number of available beds for specialty 1, as reported in Table~\ref{tab:bedsAinput} for the plot at the top, Table~\ref{tab:bedsBinput} for the middle one, and Table~\ref{tab:bedsCinput} for the bottom one, for each day of the week, from Monday through Friday. The colored part of the bars indicates the amount of occupied beds while the gray part the beds left unoccupied by our planning.


\subsection{Scalability Analysis}

We have performed a scalability analysis on the performance of employed ASP solver and encoding for ORS with bed management w.r.t schedule length. 

\paragraph{\bf Evaluation.} The characteristics of the tests for each scenario are the following:
\begin{itemize}
\item We consider 7 different benchmarks with planning period of 1, 2, 3, 5, 7, 10, and 15 working days;
\item For each benchmark the total number of randomly generated  registrations were 70 per day, i.e. 70, 140, 210, 350, 490, 700, and 1050 for 1, 2, 3, 5, 7, 10, and 15 days, respectively;
\item 5 specialties for each benchmark;
\item 10 ORs unevenly distributed among the specialties;
\item 5 hours long morning and afternoon shifts for each OR summing to 100, 200, 300, 500, 700, 1000, and 1500 hours of OR available time for the 7 benchmarks;
\item An execution time of 60 seconds was given to each instance.
    \end{itemize}
Table~\ref{tab:reg} shows the distribution of the total number of randomly generated registrations for each benchmark. Further, this table also shows the distribution of ORs for each speciality, i.e to speciality 1 three ORs are assigned, to speciality 2, 3 and 5, two ORs are assigned, while to speciality 4 only one OR is allocated. Each speciality is considered as a ward with variable number of available beds.

\begin{table}[ht]
    \caption{Total number of randomly generated registrations for each benchmark}
	\centering
	\label{tab:reg}
	\begin{tabular}{ccccccccc}
	\hline\hline
		Specialty & \multicolumn{7}{c}{Registrations} & ORs \\
		\cline{2-8}\\
		& 15-day & 10-day & 7-day	& 5-day   & 3-day  & 2-day  & 1-day  \\ \hline
		1               & 240  & 160     & 112   & 80      & 48     & 32     & 16     & 3                                                                                                                               \\
		2               & 210  & 140     & 98    & 70      & 42     & 28     & 14     & 2                                                                                                                                \\
		3               & 210  & 140     & 98    & 70      & 42     & 28     & 14     & 2                                                                                                                                \\
		4               & 180  & 120     & 84    & 60      & 36     & 24     & 12     & 1                                                                                                                                 \\
		5               & 210  & 140     & 98    & 70      & 42     & 28     & 14     & 2                                         \\
		\hline
		Total           & 1050 & 700     & 490   & 350     & 210    & 140    & 70     & 10      \\
		\hline\hline
	\end{tabular}
\end{table}

Moreover, we kept also for this analysis the parameters for the random generation of scheduler input from Table~\ref{tab:schedinput}.


\paragraph{\bf Scenario A.}
The results of scenario A are reported in Table~\ref{ScalabilityA}, that shows averages of results for satisfiable instances with abundance of available beds. It should be noted that each benchmark (15, 10, 7, 5, 3, 2 and 1 day) represents here the average of the satisfiable runs with different randomly generated inputs. 

As we can see, Table~\ref{ScalabilityA} contains seven columns, where the first column shows the name of the benchmark for which the test is performed, the columns from the second to the fourth show the average number of the assigned registrations out of the generated ones for each priority (P1, P2 and P3, respectively), while the last two columns show the mean for the OR time and the bed occupancy efficiency. 

As we can observe from the results, all considered benchmarks achieved an overall OR time efficiency greater than 80\%, but for the case with 15 days, where some degradation is visible, for which the OR time Efficiency is 67.6\%. 

\begin{table}[ht]
    \caption{Averages of the results for 15, 10, 7, 5, 3, 2 and 1 day benchmarks for Scenario A.}
    \centering
    \label{ScalabilityA}
    \begin{tabular}{ccccccc}
    \hline\hline
    Benchmark & Priority 1 & Priority 2 & Priority 3 & Total     & 
    OR time Eff. & Bed Occupancy Eff.\\
    \hline
    15 days & 211 / 211 & 261 / 421 & 70 / 418 & 542 / 1050 & 67.6\% & 62.5\% \\
    10 days & 142 / 142 & 240 / 277 & 76 / 281 & 458 / 700 & 84.4\% & 61.6\% \\
    7 days  & 105 / 105 & 166 / 189 & 68 / 196 & 339 / 490 & 89.8\% & 53.3\% \\
    5 days & 68 / 68 & 131 / 139 & 65 / 143 & 264 / 350 & 96.2\% & 51.6\% \\
    3 days  & 44 / 44   & 78 / 82   & 36 / 84  & 158 / 210 & 96.5\% & 36.9\% \\
    2 days  & 28 / 28   & 53 / 57   & 22 / 55  & 108 / 140 & 95.9\% & 27.6\% \\
    1 day   & 13 / 13   & 27 / 29   & 13 / 28  & 53 / 70   & 93.4\% & 15.3\% \\
    \hline\hline
    \end{tabular}
\end{table}

\paragraph{\bf Scenario B.} The results of the scalability analysis for scenario B are reported in Table~\ref{ScalabilityB} that shows averages of results for all satisfiable instances generated, and is organized as Table~\ref{ScalabilityA}. 


From Table~\ref{ScalabilityB} it can be observed that all the considered benchmarks achieve efficiency of bed occupancy greater than 90\%, in particular between 92\% and 96\%, being able, as in the previous scenario, to optimize the constrained resource.

\begin{table}[ht]
    \caption{Averages of the results for 15, 10, 7, 5, 3, 2 and 1 day benchmarks for Scenario B.}
    \centering
    \label{ScalabilityB}
    \begin{tabular}{ccccccc}
    \hline\hline
    Benchmark & Priority 1 & Priority 2 & Priority 3 & Total & 
       OR time Eff. & Bed Occupancy Eff.\\
    \hline
    15 days & 208 / 208 & 189 / 423 & 107 / 419 & 504 / 1050 & 62.7\% & 96.9\%\\
    10 days & 144 / 144 & 166 / 281 & 56 / 275  & 366 / 700 & 68.0\% & 96.5\% \\
    7 days  & 95 / 95   & 136 / 197 & 30 / 198  & 261 / 490 & 70.6\% & 94.0\% \\
    5 days & 68 / 68 & 88 / 139 & 31 / 143 & 187 / 350 & 68.8\% & 94.0\% \\
    3 days  & 41 / 41   & 58 / 85   & 19 / 84   & 118 / 210 & 73.3\% & 93.3\% \\
    2 days  & 30 / 30   & 35 / 56   & 6 / 54    & 71 / 140  & 66.7\% & 92.6\% \\
    1 day   & 16 / 16   & 25 / 28   & 6 / 26    & 47 / 70   & 70.4\% & 99.4\% \\
    \hline\hline
\end{tabular}
\end{table}

\paragraph{\bf Scenario C.}
The results for third scenario are reported in Table~\ref{ScalabilityC}, which is organized as Tables~\ref{ScalabilityA} and~\ref{ScalabilityB}.



From the results, we can note that although the total number of available beds if further reduced, for all benchmarks we achieve efficiency of bed occupancy greater than 85\%, even for the extreme case of 15 days planning length, and overall between 86\% and 95\%.

\begin{table}[ht]
    \caption{Averages of the results for 15, 10, 7, 5, 3, 2 and 1 day benchmarks for Scenario C.}
    \centering
    \label{ScalabilityC}
    \begin{tabular}{ccccccc}
    \hline \hline
    Benchmark & Priority 1 & Priority 2 & Priority 3 & Total & 
       OR time Eff. & Bed Occupancy Eff.\\
    \hline
    15 days & 206 / 206 & 64 / 407 & 32 / 437 & 302 / 1050 & 37.8\% & 96.4\% \\
    10 days & 135 / 135 & 91 / 289 & 13 / 276  & 239 / 700 & 44.0\% & 96.3\% \\
    7 days  & 101 / 101 & 49 / 194 & 13 / 195  & 163 / 490 & 43.2\% & 91.6\% \\
    5 days  & 68 / 68   & 41 / 140 & 10 / 143  & 119 / 350 & 44.5\% & 91.9\% \\
    3 days  & 43 / 43   & 25 / 83  & 4 / 84    & 72 / 210 & 45.0\% & 91.4\% \\
    2 days  & 27 / 27   & 19 / 57  & 3 / 56    & 49 / 140 & 43.9\% & 85.9\% \\
    1 day   & 13 / 13   & 22 / 29   & 3 / 28   & 38 / 70  & 70.4\% & 99.4\% \\
    \hline\hline
    \end{tabular}
\end{table}

\section{ASP Encoding for the Rescheduling problem}
\label{sec:resch}

The rescheduling procedure is applied to a previously planned schedule, i.e. we start from an already created schedule (old schedule or x-schedule) that could not be executed fully till the end due to some reasons, e.g. some patients could not be operated in their assigned  slots or the patients may delete their registration. In such a situation all those postponed registrations (or surgeries) must be reallocated to one of the next slots in the remaining part of the original planning period (new schedule or y-schedule). The planning period we consider is 5 days. 


Once planned, a speciality schedule does not influence other specialties so it makes sense to reschedule one specialty at a time. Since we already have the initial schedule for the planning period of 5 days, we assumed that in day 2, a number of registrations from specialty 1 had to be postponed to the next day. So we  have to reschedule these registrations 
in the remaining available three days, i.e. day 3, 4 and 5.

In order to insert the postponed registrations in the new schedule (y-schedule) we have to make sure that the start of the schedule leaves enough available OR time by automatically dropping the necessary registrations from the old schedule; the choice of the registrations to be removed will begin from the last day, i.e. day 5 of the planning period and from registrations in the priority 3 category.


The next two sub-sections will show ASP encoding for the rescheduling problem, together with the needed changes in the data model, and the results of the experimental analysis we performed.

\subsection{ASP Encoding}
\begin{figure*}[t!]
\begin{asp}
$(rr_{1})$$\phantom{_0}$ {y(R,P,O,S,D)} :- registration(R,P,_,_,SP,_,_), mss(O,S,SP,D).
$(rr_{2})$$\phantom{_0}$ :- y(R,P,O,S1,_), y(R,P,O,S2,_), S1 != S2.
$(rr_{3})$$\phantom{_0}$ :- y(R,P,O1,S,_), y(R,P,O2,S,_), O1 != O2.
$(rr_{4})$$\phantom{_0}$ surgery(R,SU,O,S) :- y(R,_,O,S,_), registration(R,_,SU,_,_,_,_).
$(rr_{5})$$\phantom{_0}$ :- y(_,_,O,S,_), #sum{SU,R: surgery(R,SU,O,S)}>N, duration(N,O,S).
$(rr_{6})$$\phantom{_0}$ stay(R,(D-A)..(D-1),SP) :- registration(R,_,_,LOS,SP,_,A), 
$\qquad\qquad$y(R,_,_,_,D), A>0.
$(rr_{7})$$\phantom{_0}$ stay(R,(D+ICU)..(D+LOS-1),SP) :- registration(R,_,_,LOS,SP,ICU,_), 
$\qquad\qquad$y(R,_,_,_,D), LOS>ICU.
$(rr_{8})$$\phantom{_0}$ stayICU(R,D..(D+ICU-1)) :- registration(R,_,_,_,_,ICU,_), 
$\qquad\qquad$y(R,_,_,_,D), ICU>0.
$(rr_{9})$$\phantom{_0}$ :- #count{R: stay(R,D,SP)} > AV, SP>0, beds(SP,AV,D).
$(rr_{10})$ :- #count{R: stayICU(R,D)} > AV, beds(0,AV,D).
$(rr_{11})$ :- not y(R,P,_,_,_),x(R,P,_,_,2).
$(rr_{12})$ totReg(L,4) :- M = #count{R:y(R,1..2,_,_,_),x(R,1..2,_,_,_)}, 
$\qquad\qquad$P = #count{R : x(R,1..2,_,_,_)}, L=P-M.
$(rr_{13})$ totReg(L,3) :- M = #count{R:y(R,3,_,_,_),x(R,3,_,_,3..4)}, 
$\qquad\qquad$P = #count{R : x(R,3,_,_,3..4)}, L=P-M.
$(rr_{14})$ totReg(L,2) :- M = #count{R:y(R,3,_,_,5),x(R,3,_,_,5)}, 
$\qquad\qquad$P = #count{R : x(R,3,_,_,5)}, L=P-M.
$(rr_{15})$ :~ totReg(L,V). [L@V]
$(rr_{16})$ :~ y(R,_,_,_,D),x(R,_,_,_,OldD),DF = |D - OldD|. [DF@1, R]
\end{asp}
\caption{ASP encoding of the ORS problem without beds-management \textit{(Rescheduling)}}
    \label{fig:encodingrescheduling}
\end{figure*}

\paragraph{Input.}
The input data is specified by means all of the atoms described in Section~\ref{sec:datamodelors} and by atoms of the form \textit{x(R,P,O,S,D)}. The latter represent a solution to the ORS problem as computed by the encoding described in Section~\ref{sec:encodingors}.
Moreover, in the following we assume that atoms of the form \textit{mss(O,S,SP,D)} include only elements from day 3 to day 5, i.e. $3 \leq D \leq 5$.

\paragraph{Output.}
The output of the new schedule is represented by atoms of the form \textit{y(R,P,O,S,D)}, representing that the registration \textit{R} of the patient with priority \textit{P} is assigned to an operating room \textit{O} in a shift \textit{S} of day \textit{D}.

\paragraph{Encoding.} 
The ASP rescheduling encoding for ORS problem is shown in Figure~\ref{fig:encodingrescheduling}. 
In particular, rules ($rr_1$)-($rr_{10}$) correspond to rules ($r_1$)-($r_{10}$) from the encoding reported in Figure~\ref{fig:encoding1}, where atoms over the predicate $x$ are replaced with the ones over the predicate $y$.
Rule ($rr_{11}$) states that all registration scheduled for the day 2 (i.e. the ones postponed according to our scenarios) should be rescheduled.
Rules ($rr_{12}$)-($rr_{15}$) ensure that the maximum number of registrations from the old schedule should be included also in the new one. In particular, rule ($rr_{12}$) computes the difference between the total number of registrations with priority 1 and 2 assigned in the previous schedule and the number of registrations assigned in the current schedule, whereas rule ($rr_{13}$) (resp. ($rr_{14}$)) computes the difference between the total number of registrations with priority 3 for day 3 and 4 (resp. day 5), and the number of registrations assigned in the current schedule.
Such differences are then minimized by means of the weak constraint ($rr_{15}$).
Finally, rule ($rr_{16}$) minimizes the total sum of the difference (in terms of number of days) between the new schedule and the old one for each registration. 

\subsection{Experimental Results}
The results using our ASP rescheduling encoding on four scenarios (I, II, III, and IV) are summarized in Table \ref{tab:re-schedulerresult}. An execution time of 60 seconds was given for each scenario. 

For our tests we started from an old schedule (x-schedule) calculated under the conditions delineated in Scenario A of the scheduling problem, in particular we took into account the results for specialty 1. Since we consider postponed registrations from a single speciality for our analysis, it should be noted that the total number of old registrations of specialty 1 with priority 1, 2 and 3 to be rescheduled in the next 3 days were 45. In the table, the first column mentions the scenario, the second shows the number of registrations that were inserted in each scenario (Postponed Registrations), the third column reports the total number of registrations from the old schedule (Total Old Registrations), while the fourth column shows the necessary number of dropped registrations from the old schedule. Finally, the last column shows the total number of registrations in the new schedule by also reassigning the postponed registrations (Total New Registrations) in the next days. Columns from the second to the fifth report the (rounded) mean of the 10 instances. Overall, our encoding dropped 0, 1, 2 and 4 priority 3 old schedule registrations for scenarios I, II, III, and IV, respectively. 

\begin{table}[t]
	\centering
	\caption{Results for the four rescheduling scenarios}
	\label{tab:re-schedulerresult}
	\begin{tabular}{ccccc}
		\hline\hline
		Scenario & Postponed Registr. & Total Old Registr.&
		Dropped Registr. &		Total New Registr. \\
\hline
		I        & 1     & 45   & 0  & 46 \\
		II        & 2     & 44   & 1  & 46 \\
		III        & 4     & 42   & 2  & 46 \\ 
		IV        & 6     & 40   & 4	 & 46 \\\hline\hline
	\end{tabular}
\end{table}

These results confirm that we managed to produce a new schedule in case of disruption of the previous one even when already in its execution phase. This was accomplished by allowing the rescheduled registrations to change time but minimizing the number of changes of surgery date, which would imply a major disruption in the procedure of the hospital, in particular regarding the ICU and ward bed management. On the one hand, in a situation of minor disruption like that of Scenario I we managed to produce a new schedule without having to drop any of the previously scheduled surgeries; on the other hand, even in case of greater disruption, like in Scenario II, III, and IV, the number of scheduled surgeries in the new schedule were at least as many as in the previous schedule. 

\section{Related Work}
\label{sec:rel}

This paper is an extended and revised version of \cite{DBLP:conf/ruleml/DodaroGKMP19} having the following main additions: $(i)$ a scalability analysis of our solution w.r.t. schedule length, $(ii)$ a rescheduling ASP encoding and its evaluation on some scenarios, and $(iii)$ a web framework for managing the experiments performed in this paper. Moreover, the analysis about the scheduling part has been performed on the third scenario where the number of beds is extremely scarce.

In this section we review related literature, organized into two paragraphs. The first paragraph is devoted to outlining different techniques for solving the ORS problem, with focus on the inclusion of bed management, while in the second paragraph we report about other scheduling problems where ASP has been employed. 

\paragraph{\bf Solving ORS problems.} Aringhieri et al. \cite{aringhieri_two_2015} addressed the joint OR planning (MSS) and scheduling problem, described as the allocation of OR time blocks to specialties together with the subsets of patients to be scheduled within each time block over a one week planning horizon. They developed a 0-1 linear programming formulation of the problem and used a two-level meta-heuristic to solve it. 
Its effectiveness was demonstrated through numerical experiments carried out on a set of instances based on real data and resulted, for benchmarks of 80-100 assigned registrations, in a 95-98\% average OR utilization rate, for a number of ORs ranging from 4 to 8. The execution times were around 30-40 seconds.
In \cite{landa_hybrid_2016}, the same authors introduced a hybrid two-phase optimization algorithm which exploits
neighborhood search techniques combined with Monte Carlo simulation, in order to solve the joint advance and allocation scheduling problem, taking into account the inherent uncertainty of surgery durations. In both the previous works, the authors solve the bed management part of the problem limited to weekend beds, while assuming that each specialty has its own post-surgery beds from Monday to Friday with no availability restriction.
In \cite{ARINGHIERI2015173}, some of the previous authors face the bed management problem for all the days of the week, with the aim to level the post-surgery ward bed occupancies during the days, using a Variable Neighbourhood Search approach.

Other relevant approaches are: Abedini et al. \cite{abedini_operating_2016}, that developed a bin packing model with a multi-step approach and a priority-type-duration rule;
Molina-Pariente et al. \cite{molina-pariente_new_2015}, that tackled the problem of assigning an intervention
date and an OR to a set of surgeries on the waiting list, minimizing the access time for patients
with diverse clinical priority values; and
Zhang et al. \cite{zhang_stochastic_2017}, that addressed the problem of OR planning with different demands from both elective patients and non-elective ones, 
with priorities in accordance with urgency levels and waiting times. However, bed management is not considered in these three last mentioned approaches.

\paragraph{\bf ASP in scheduling problems.} We already mentioned in the introduction that ASP has been already successfully used for solving hard combinatorial and application problems in several research areas. Concerning ORS, the problem has been already addressed in \cite{DBLP:conf/aiia/DodaroGMP18,DBLP:journals/ia/DodaroGMP19}, but without taking into account beds, which instead are a fundamental resource to be considered. 
Concerning scheduling problems other than ORS, ASP encodings were proposed for the following problems:
	\textit{Incremental Scheduling Problem}~\cite{DBLP:conf/lpnmr/Balduccini11,DBLP:journals/ai/CalimeriGMR16,DBLP:journals/jair/GebserMR17,DBLP:conf/lpnmr/GebserMR17}, where the goal is to assign jobs to devices such
	that their executions do not overlap one another;
	\textit{Team Building Problem}~\cite{DBLP:journals/tplp/RiccaGAMLIL12}, where the goal is to allocate the available personnel of a seaport for serving the incoming ships;
	\textit{Nurse Scheduling Problem}~\cite{DBLP:conf/aiia/AlvianoDM17,DBLP:conf/lpnmr/DodaroM17,DBLP:journals/ia/AlvianoDM18}, where the goal is to create a scheduling for nurses working in hospital units;
	\textit{Interdependent Scheduling Games}~\cite{DBLP:conf/aiia/Amendola18a}, which requires interdependent services among players, that control only a limited number of services and schedule independently, and the \textit{Conference Paper
Assignment Problem}~\cite{DBLP:conf/aiia/AmendolaDLR16}, which deals with the problem of assigning reviewers in the Programme Committee to submitted conference papers. Other relevant papers are Gebser et. al~\cite{DBLP:journals/tplp/GebserOSR18}, where, in the context of routing driverless transport vehicles,  the setup problem of routes such that a collection of transport tasks is accomplished in case of multiple vehicles sharing the same operation area is solved via ASP, in the context of car assembly at Mercedes-Benz Ludwigsfelde GmbH, and the recent survey paper by Falkner et al.~\cite{DBLP:journals/ki/FalknerFSTT18}, where industrial applications dealt with ASP are presented, including those involving scheduling problems.



\section{Web Framework}
\label{sec:appl}
\begin{figure}
    \centering
    \scalebox{0.33}{\includegraphics{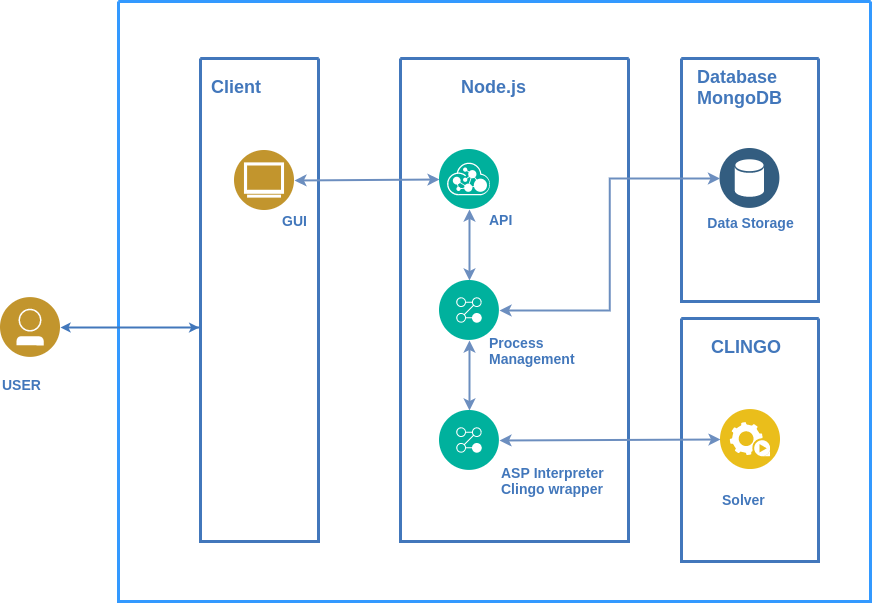}}
    \caption{Web application architecture schema.}
    \label{fig:GUI-architecture}
\end{figure}

\begin{figure}
    \centering
    \scalebox{0.20}{\includegraphics{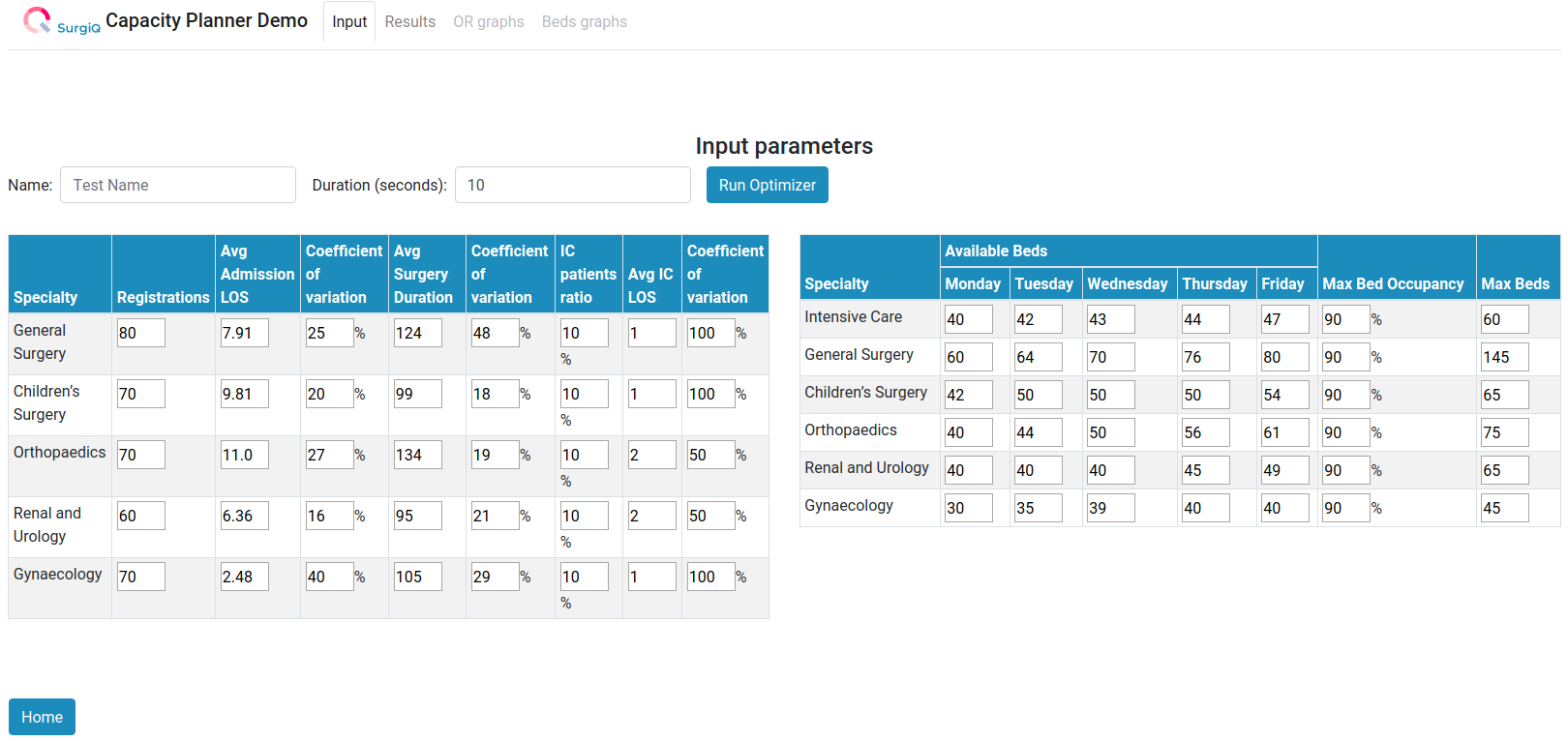}}
    \caption{Input screen for the registration generator parameter and the beds availability.}
    \label{fig:GUI-data-insert}
\end{figure}

Our scheduling solution has been planned and developed also in view of a practical utilization by medical operators in hospitals. To this end, we are developing a web application which wraps the ASP encoding and the {\sc clingo} solver (see Figure \ref{fig:GUI-architecture}).
The software is a full-stack JavaScript application with a Graphical User Interface (GUI) and a Node.js back-end. 
The ASP facts and encoding are dynamically composed at run-time reflecting the user choices, and are then relayed to the ASP solver through a wrapper package.
This solution allows the solver to be embedded inside an easily reachable and usable web application, removing the hurdle that installing and managing the solver may represent for a non-expert user.
The application currently includes:
\begin{itemize}
    \item a registration and authentication process,
    \item a database for storing and retrieving previous test data, and
    \item a GUI to easily create and customize new test scenarios or load pre-made ones.
\end{itemize}
The GUI can be divided in the following sections: an input screen, an overall results screen and their graphical representation for the ORs and for the bed occupancy.
The input screen (see Figure \ref{fig:GUI-data-insert}) currently hosts two tables: on the left the parameters for the random generation of the registrations and on the right the bed availability. It is important to note that in a more operative stage the generation parameter will be obviously replaced by the actual registration data.
In the left table the user can set the parameters for the generator. From left to right these are:
\begin{itemize}
    \item the specialty names;
    \item the number of registrations we aim to assign for each specialty;
    \item the parameters (mean and coefficient of variation) of the Gaussian distribution used to generate the predicted LOS in the ward after the surgery; 
    \item the parameters (mean and coefficient of variation) of the Gaussian distribution used to generate the predicted surgery lengths;
    \item the ratio of patients predicted to need a place in the ICU ward; and
    \item the parameters (mean and coefficient of variation) of the Gaussian distribution used to generate the predicted LOS in the ICU ward after the surgery. 
\end{itemize}
In the right table, the user can set the number of available beds for each ward connected to a specialty. From left to right the parameters are:
\begin{itemize}
    \item the specialty names;
    \item the number of available beds for each day of the planning period (in the figure a five days period is shown);
    \item the percentage of the maximum number of beds potentially available, reflecting the practice usually used by hospitals to reserve a bed quota for emergencies and unexpected events; and
    \item the total number of beds allocated to the specialty.
\end{itemize}
The number of beds tends to increase during the planning period to simulate the patients operated during the previous period that still occupy a bed at the beginning of the planning period and are gradually discharged.

\begin{figure}
    \centering
    \scalebox{0.20}{\includegraphics{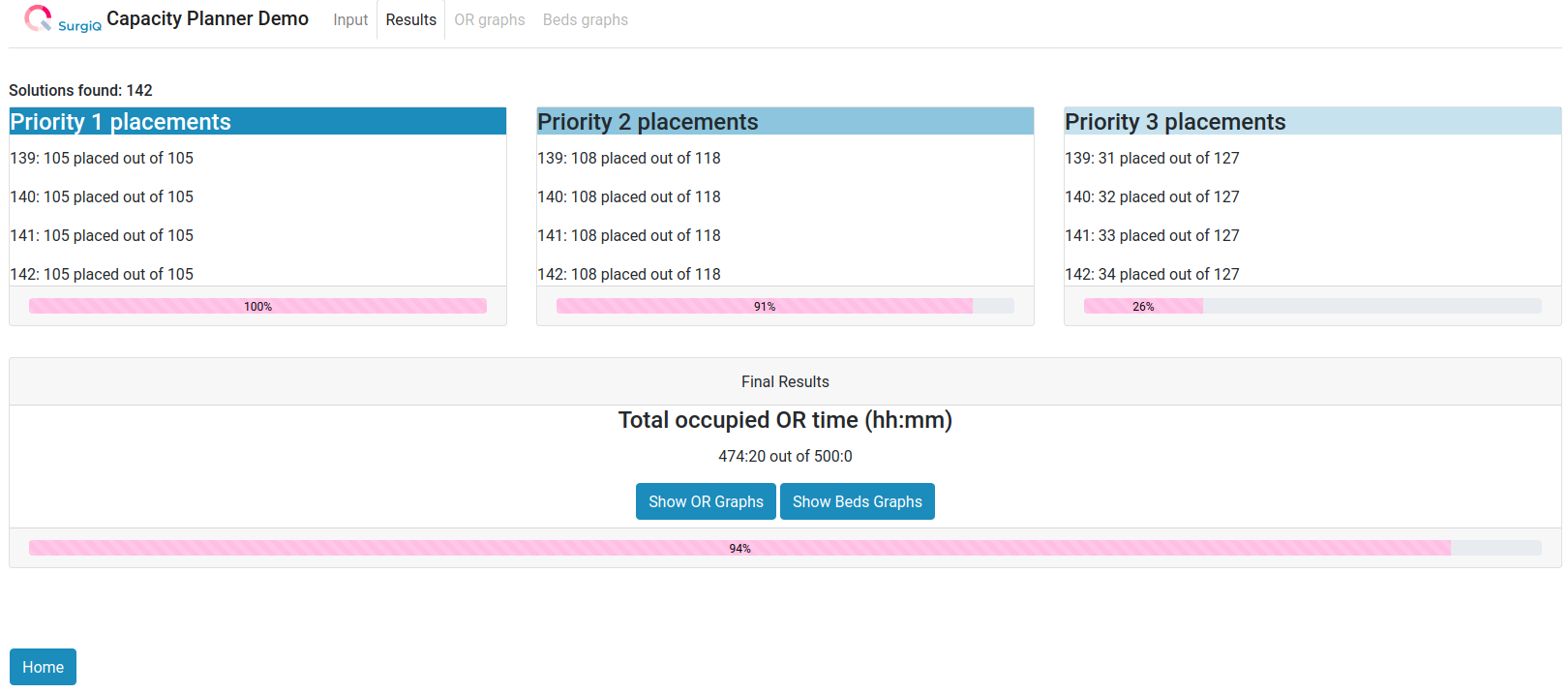}}
    \caption{Results screen.}
    \label{fig:GUI-results}
\end{figure}

In the overall results screen (see Figure \ref{fig:GUI-results}) the user can monitor in real-time the evolution of the process and, finally, read the final results:
\begin{itemize}
    \item At the top of the screen there are three cards containing the number of assigned registrations out of the total, arranged according to their priority class, for each solution found by the solver engine. Each number is continuously updated during the execution whenever a new solution is found. The percentage of assigned registrations is represented by a progress bar at the bottom of each card.
    \item At the bottom we summarize the final results at the end of the execution. In particular, the OR time out of the total available is reported, both in absolute numbers and as a percentage through a progress bar.
    \item Finally, we have two links that lead to the graphical representation of the OR scheduling and bed occupancy for each day of the planning period.
\end{itemize}

\begin{figure}
    \centering
    \scalebox{0.25}{\includegraphics{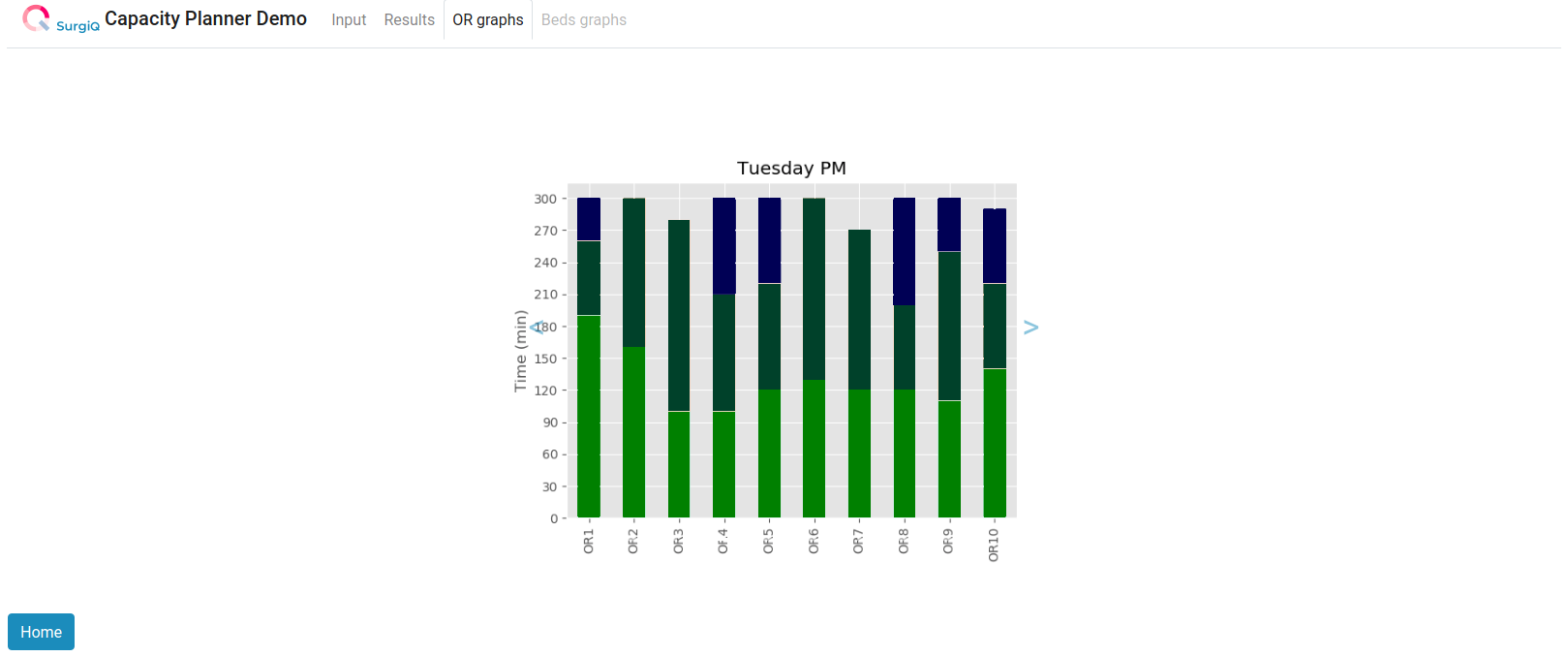}}
    \caption{Graphical representation of the OR schedules for a single day and shift.}
    \label{fig:GUI-OR-schedule}
\end{figure}

\begin{figure}
    \centering
    \scalebox{0.25}{\includegraphics{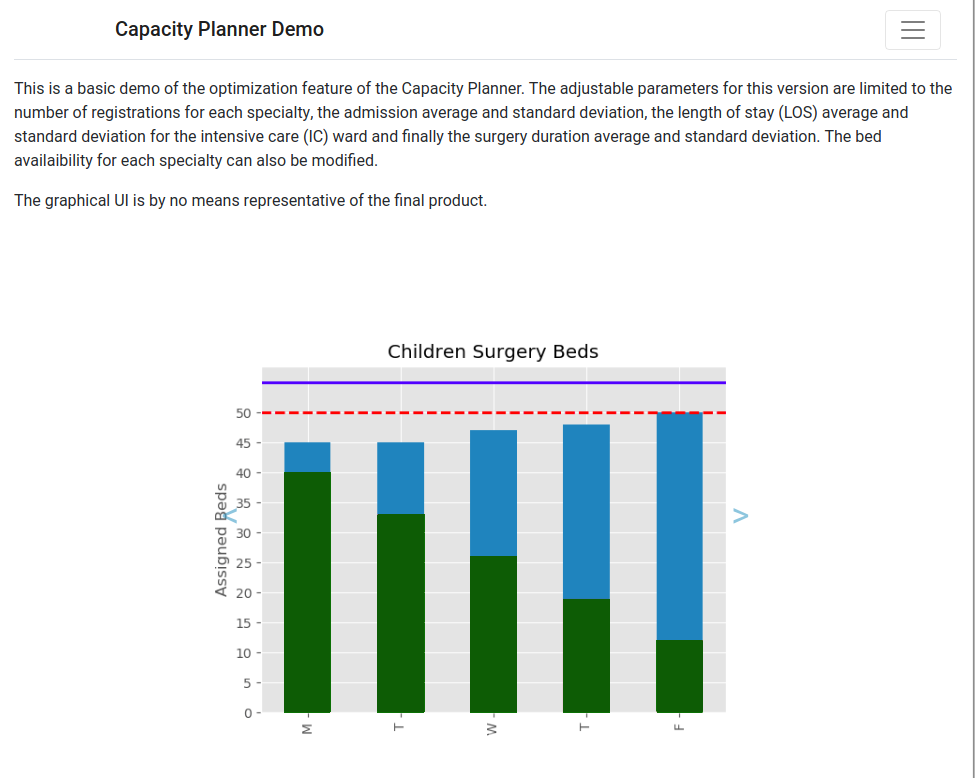}}
    \caption{Graphical representation of the bed occupancy for a single specialty.}
    \label{fig:GUI-beds-occupancy}
\end{figure}

The graphical representation of each OR schedule is shown through colored bars, one for each day and shift of the period (see Figure \ref{fig:GUI-OR-schedule}). In the OR graphs each column displays an OR and each bar inside the column represents a registration. 
The bed occupancy is also shown through a carousel of stacked bar graphs (see Figure \ref{fig:GUI-beds-occupancy}). This time each graph displays the bed occupancy of a single specialty: each column shows the situation in a day, in particular the green part of the bar denotes the beds already occupied by patients operated previously, while the blue part shows the beds occupied by the patients scheduled in the current period. The solid line shows the total number of beds assigned to the specialty, while the dashed line gives the maximum number of available beds, respecting the quota for emergencies.
\section{Conclusions}
\label{sec:conc}
In this paper we have employed ASP for solving the ORS problem with bed management, given ASP has already proved to be a viable tool for solving scheduling problems due to the readability of the encoding and availability of efficient solvers. Specifications of the problem are modularly expressed as rules in the ASP encoding, and the ASP solver {\sc clingo} has been used.
We have then presented the results of experimental and scalability analysis, on ORS benchmarks with realistic sizes and parameters on three scenarios, that reveal that our solution is able to utilize efficiently whichever resource is more constrained, being either the OR time or the beds. Moreover, for the planning length of 5 days usually used in small-medium Italian hospitals, this is obtained in short timings in line with the needs of the application. We further developed and evaluated a rescheduling procedure, to be employed in case the original schedule cannot be fully executed for some reason. We finally also presented a web framework that supports the online execution of our scheduling solution. While we do not directly manage emergencies, the flexibility of our algorithm can be exploited even in those cases. For example, if a part of the hospital resources (being OR time, ICU or ward beds) must be suddenly redirected to serve other purposes, as for example happened in the Covid19 emergency, by simply adjusting the numbers our solution can immediately be utilized to manage the remaining resources at the best of their capacity.  
Future work includes the usage of other solving paradigms as, e.g. SMT or CSP, and the extension of our solution for including, e.g. surgical teams and/or the Post Anesthesia Care Unit. In terms of efficiency, we plan both to evaluate heuristics and optimization techniques (see, e.g.~\cite{DBLP:conf/jelia/GiunchigliaMT02,DBLP:conf/cp/GiunchigliaMT03,DBLP:conf/ecai/RosaGM08,journals/jlc/AlvianoDMR15}), as well as further clingo options, and improvement to the current encoding. In this light, we have already performed some preliminary analysis on the (costly) rules $(r_{11})$-$(r_{13})$ of our encoding, substituted with an equivalent formulation that avoids the \#count aggregate. Preliminary tests on our biggest, i.e. 15 days, instances show that the alternative formulation may lead to advantages.  

\noindent
All materials presented in this work, including benchmarks and encodings, can be found at:
\url{http://www.star.dist.unige.it/~marco/RuleMLRR2TPLP/material.zip}. 

\paragraph{\bf Acknowledgments.} 

The research of three of the authors
of the paper, Ivan Porro, Giuseppe Galat\`a and  Muhammad Kamran Khan, is partially funded by the "POR FESR Liguria 2014-2020" public grant scheme, and by EIT Health through the Headstart 2020 grant received by SurgiQ srl  (Reference: 2020 Headstart Programme, PoC activity \#20206, Innostars, 2020-HS-0339).

\bibliographystyle{acmtrans}
\bibliography{main}

\end{document}